\documentclass[conference]{IEEEtran}
\usepackage{url}
\usepackage{algorithm}
\usepackage{algorithmic}
\usepackage{ upgreek }
\ifCLASSINFOpdf
\usepackage[pdftex]{graphicx}
\else
\usepackage[dvips]{graphicx}
\fi
\usepackage{amsmath}

\usepackage{amsfonts} 

\usepackage{color}
\usepackage{ dsfont }
\usepackage[misc]{ifsym}
\usepackage{algorithmic}
\usepackage{array}
\usepackage{verbatim}
\ifCLASSOPTIONcompsoc
\usepackage[caption=false,font=normalsize,labelfont=sf,textfont=sf]{subfig}
\else
\usepackage[caption=false,font=footnotesize]{subfig}
\fi
\usepackage{amsmath}

\begin{document}

\title{Vulnerability Analysis for Data Driven\\ Pricing Schemes}

\author{\IEEEauthorblockN{Jingshi Cui, Haoxiang Wang, Chenye Wu, \emph{and} Yang Yu$^\text{\Letter}$}
\IEEEauthorblockA{Institute for Interdisciplinary Information Sciences\\
Tsinghua University, Beijing, 100084, P.R. China\\
Email:yangyu1@tsinghua.edu.cn}
}

\maketitle

\begin{abstract}
Data analytics and machine learning techniques are being rapidly adopted into the power system, including power system control as well as electricity market design. In this paper, from an adversarial machine learning point of view, we examine the vulnerability of data-driven electricity market design. More precisely, we follow the idea that consumer's load profile should uniquely determine its electricity rate, which yields a clustering oriented pricing scheme. We first identify the strategic behaviors of malicious users by defining a notion of disguising. Based on this notion, we characterize the sensitivity zones to evaluate the percentage of malicious users in each cluster. Based on a thorough cost benefit analysis, we conclude with the vulnerability analysis.
\end{abstract}

\vspace{0.1cm}
\begin{IEEEkeywords}
Electricity Market, Adversarial Machine Learning, Sensitivity Analysis
\end{IEEEkeywords}

\IEEEpeerreviewmaketitle

\section{Introduction \protect \footnote{This work has been supported in part by the Youth Program of National Natural Science Foundation of China (No. 71804087), National Key R\&D Program of China (2018YFC0809400), Turing AI Institute of Nanjing, and Zhongguancun Haihua Institute for Frontier Information Technology.}}\label{Introduction}
The wide deployment of smart meters boosts the data analytics' penetration in power system operation, which is reshaping the electricity economy's landscape. However, the technological limits in data analytics generate a sequence of new risks, threatening the power system reliability, security, and economic efficiency.

\subsection{Opportunities and Challenges}
The current primary impediment on widely utilizing data analytics methods comes from the privacy concern. Such concern is diminishing with the advance in privacy preserving computation. In contrast, it has not attracted sufficient attention on whether data analytics may generate new market loopholes that can be utilized for market manipulation. In other words, while more information from the demand side is able to improve the market efficiency, it remains unknown if there will be any strategic behaviors with a direct implementation of machine learning algorithms in electricity market.

In this paper, we submit that utilizing the demand data can lead to both efficiency improvement opportunity and new market manipulation loophole. The detailed demand data enables the system operator to use $k$-means clustering algorithm to assess each customer's marginal system impact, which serves as the basis for a data-driven pricing scheme \cite{Yang2017Good}. However, due to the structure of resulted clusters, we identify the existence of market manipulation.

%We follow the research by Yu $\textit{et al.}$ in \cite{Yang2017Good}, where the authors prove that the demand profile should uniquely determine each consumer's electricity price. Using the notion of clustering, Yu $\textit{et al.}$ propose the most efficient pricing scheme. We seek to investigate any strategic behavior facing such a pricing scheme: we define disguising to distinguish malicious consumers from the population. Then, we characterize the sensitivity zone for each cluster to briefly quantify the vulnerability of this pricing scheme. We conclude with the vulnerability analysis via the cost benefit analysis. Figure \ref{framework} summarizes this paradigm.

\begin{figure}[!t]
\centering
\includegraphics[width=3 in]{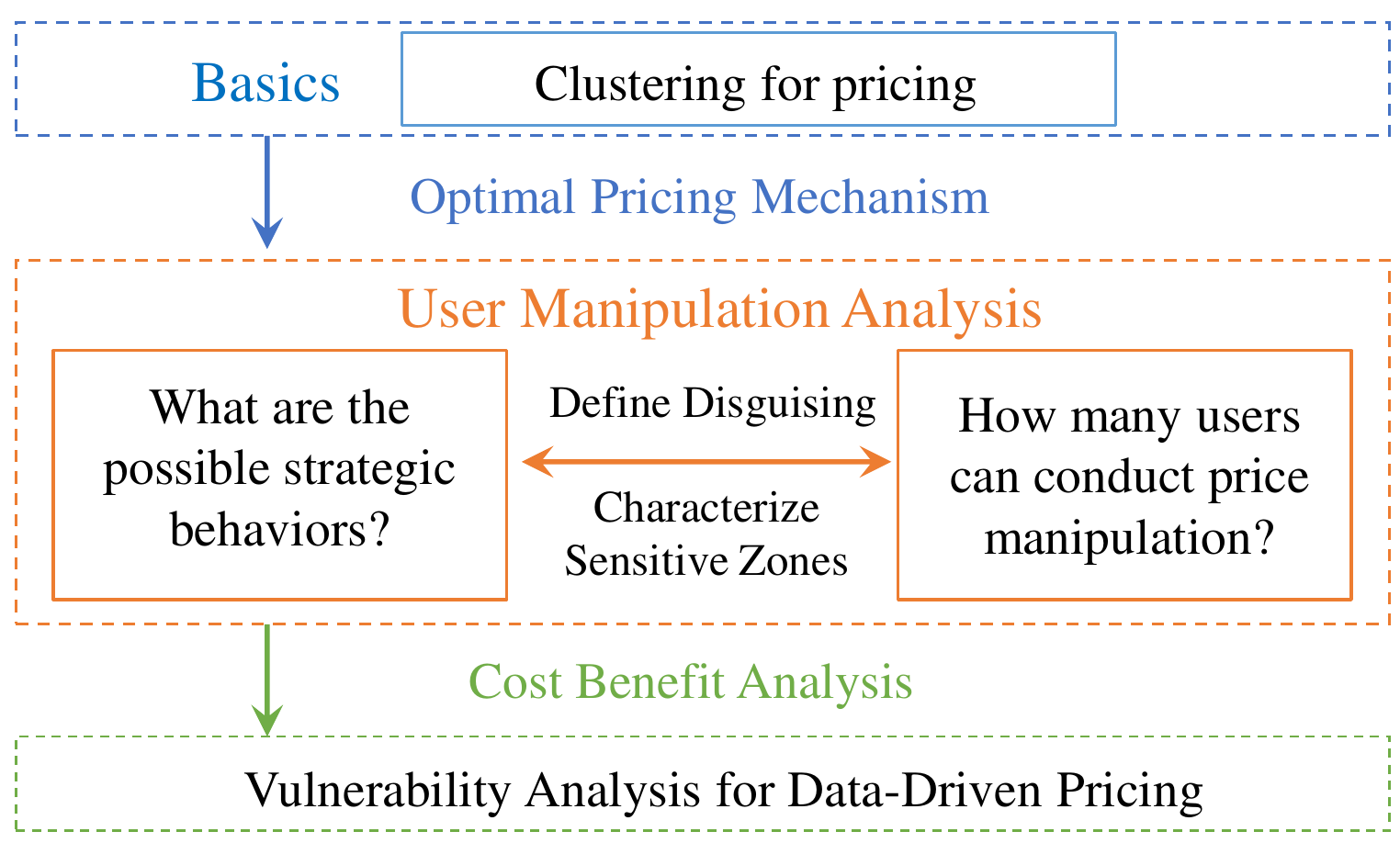}
\caption{Paradigm of Our Vulnerability Analysis}\vspace{-0.3cm}
\label{framework}
\end{figure}

\subsection{Related Work}
The major body of related literature that uses machine learning techniques for better power system operation focuses on load prediction. Just to name a few, Kong $\textit{et al.}$ exploit the temporal characteristics in residential load and propose an LSTM (Long Short-Term Memory) load forecasting method in \cite{Kong2017Short}. Many deep learning models have also been employed for more accurate load prediction, including pooling-based deep recurrent neural network in \cite{Shi2018Deep}, FFNN (Feed-Forward Neural Network) in \cite{Wei2019Prediction}, DBN (Deep Belief Network) in \cite{Dedinec2016Deep}, etc. Chen $\textit{et al.}$ seek to integrate domain knowledge in different neural network building blocks in \cite{Chen2018Short}. Researchers have also used machine learning to solve scheduling or dispatch problems for electricity market. For example, Mocanu $\textit{et al.}$ propose a deep reinforcement learning method to conceive an online optimization for building energy management systems in \cite{Mocanu2017On}.

Another line of research exploits opportunities in designing better pricing schemes for demand side. Based on utility maximization, Samadi $\textit{et al.}$ propose an optimal real-time pricing algorithm for demand side management in \cite{Samadi2010Optimal}. By combining hybrid particle swarm optimizer with mutation algorithm, Xu $\textit{et al.}$ design a data-driven pricing scheme to help the utility minimize peak demand in \cite{Xu2017Data}. Exploiting price elasticity of electricity consumption, Yu $\textit{et al.}$ develop a parametric time-utility model to obtain the optimal real time price and maximize the social welfare in \cite{Yu2012A}. Qian $\textit{et al.}$ develop an SAPC (Simulated-Annealing-based Price Control) algorithm to reduce the peak-to-average load ratio via a two-stage optimization in \cite{Li2013Demand}.

Most studies on quantifying the risk caused by data focus on the cyber security, e.g., the false data injection attack \cite{Liang2017The}, and cyber attack on the topology errors \cite{Ashok2012Cyber} or simply incorrect topology information \cite{Kim2013On}. In contrast to the literature, we seek to understand the vulnerability in data-driven pricing schemes by examining possible strategic behaviors.

%The engineering risks caused by the data analytics' penetration in the power system has been observed. Liang  $\textit{et al.}$ consider some implications for FDIAs (False Data Injection Attack) in late 2015 Ukraine Blackout event in \cite{Liang2017The}. Ashok $\textit{et al.}$ discuss the effect of intentional topology errors created on the power system by cyber attacks in \cite{Ashok2012Cyber}. Deka $\textit{et al.}$ analyse necessary and sufﬁcient conditions for feasible attacks using a novel graph-coloring in \cite{Deka2015One}. Kim $\textit{et al.}$ describe the problem of cyber attacks on a smart grid aiming at misleading the control center with incorrect topology information in \cite{Kim2013On}. Liu $\textit{et al.}$ consider the problem of cyber-attacks due to the high integration of information technologies in \cite{Xuan2016Masking}.

\subsection{Our Contribution}
 To the best of our knowledge, we are the first to conduct vulnerability analysis for data analytics application in the electricity sector. The principal contributions of this paper are highlighted as follows.

\begin{itemize}
    \item \textit{Identify Strategic Behavior}: We propose the definition of disguising to contrast malicious users out of population. This serves as basis for identifying strategic behaviors. 
    \item \textit{Characterize Sensitive Zone}: Definition of disguising characterizes the sensitive zone for each cluster. We seek to observe the impacts of different parameters of disguising on shaping the sensitive zone.
    \item \textit{Vulnerability Analysis}: Based on the sensitive zone characterization and a thorough cost benefit analysis, we conduct the vulnerability analysis for the cluster oriented pricing scheme.
\end{itemize}

The rest of the paper is organized as follows. Section \ref{Clustering} overviews the dataset and revisits the major conclusions in using clustering for pricing. Based on this clustering approach, we define disguising to identify strategic behaviors in Section \ref{Disguising}. Section \ref{Radius} further quantifies the sensitive zones for each cluster, which serves as the basis for our vulnerability analysis in Section \ref{Vulnerability}. Finally, concluding remarks and future directions are given in Section \ref{Conclusion}.

%The following contents of the paper are mainly divided into six parts. In Section \uppercase\expandafter{\romannumeral2}, we describe the problem to be studied. We demonstrate a brief introduction to the dataset and user pricing strategy in the experiment in Section \uppercase\expandafter{\romannumeral3}. Section \uppercase\expandafter{\romannumeral4} shows disguising from good consumers and defines disguising. Section \uppercase\expandafter{\romannumeral5} characterizes sensitive zone. We analyse cost benefit in Section \uppercase\expandafter{\romannumeral6}. Finally, Section \uppercase\expandafter{\romannumeral7} summarizes the paper. 

\section{Clustering for Pricing: the Basics}\label{Clustering}

A good pricing scheme in general should reflect the user's impact on the marginal system cost. This inspires Yu $\textit{et al.}$ to encode user's load profile into the pricing scheme \cite{Yang2017Good}, which yields the notion of clustering for pricing.

\subsection{Determinant of System Impact}

In power system operation, the daily system cost of producing energy is determined by the profile of system load, which is the aggregate of individual load profiles. Hence, to evaluate each individual user's marginal impact on the system cost, we only need to know the user's load profile, the system profile, and the system cost structure. In a large power system, the last two pieces of information are almost invariant to individual behaviors. Hence, the demand profile of each user is the only determinant of its marginal system cost impact.

\subsection{Profile-Based Pricing Strategy}

Based on this observation, Yu $\textit{et al.}$ define the marginal system cost impact (termed as MCI) using $l_{1}$-norm:

\begin{equation}\label{MCI}
\begin{aligned}
MCI_{i}&=\lim_{\triangle \rightarrow 0}\frac{C(\mathbf{d}+\Delta \frac{\mathbf{d}_{i}}{||\mathbf{d}_{i}||_{1}})-C(\mathbf{d})}{\triangle }\\
&=\sum_{t=1}^{24}\frac{d_{i}^{t}}{\sum_{m=1}^{24}d_{i}^{m}}\lambda _{t}, \\
\end{aligned}
\end{equation}
where $\mathbf{d}_{i}=\left \{ d_{i}^{1},...,d_{i}^{24} \right \}$ denotes the consumer $i$'s hourly demand in a day; $\mathbf{d}$ is the aggregate load profile for the system; $C(\mathbf{d})$ denotes the system total cost given profile $\mathbf{d}$, and $\mathbf{\Lambda }=\left \{ \lambda _{1},...,\lambda _{24} \right \}$ represents the real time hourly price in a day.

With the carefully chosen $l_{1}$-norm in the definition of MCI, Yu $\textit{et al.}$ submit that given the real time price $\mathbf{\Lambda}$, the optimal pricing strategy is to set each user's daily average rate to its own MCI. Hence, consumers with the same demand profile should share the same retail price. This observation establishes the basis for clustering.

\subsection{Clustering: Implementation and Challenges}

Empirical studies have shown that consumers do share similar profiles over population \cite{6945384}, which implies that the user load profiles can be clustered into limited types. The users in the same cluster can share the same retail price, which eases the pricing scheme implementation.

However, since the users in the same cluster do not share \textit{exactly the same} load profile, a data-driven pricing approach always creates loophole allowing certain users to bypass to other clusters for a better retail price. This motivates us to examine such strategic behaviors by defining disguising.

\subsection{Dataset Overview}

Before diving into the details of analyzing the strategic behaviors, we provide a brief overview on the dataset, which we will use throughout the paper. 

We use the Pecan Street data \cite{DemandData} of around 40 users (with relatively good quality of data) from May $1$ to Aug $8$, $2015$. To better represent the clustering results for a large population, we combine the daily data of all the users into one dataset to obtain a larger dataset ($3,155$ valid load profiles in total).

Different from the classical $k$-means clustering, we use $l_{1}$-norm to replace the Euclidean distance, in response to the definition of MCI. More precisely, we use $\mathbf{\widehat{d}}_{i}= (\widehat{d}_{i}^{1}...\widehat{d}_{i}^{h}...\widehat{d}_{i}^{24})\in \mathbb{R}$ to denote daily energy consumption vector of user $i$. To better highlight the observation that user's load profile uniquely determines user's price, we first normalize each user's energy consumption according to $l_{1}$-norm:
\begin{equation}\label{normalize}
    \mathbf{d}_{i}=\frac{\mathbf{\widehat{d}}_{i}}{||\mathbf{\widehat{d}}_{i}||_{1}}=\frac{\mathbf{\widehat{d}}_{i}}{\sum_{t=1}^{24}|\widehat{d}_{i}^{t}|}.
\end{equation}

\begin{figure*}[!t]
\centering
\subfloat[]{\includegraphics[width=3.8in]{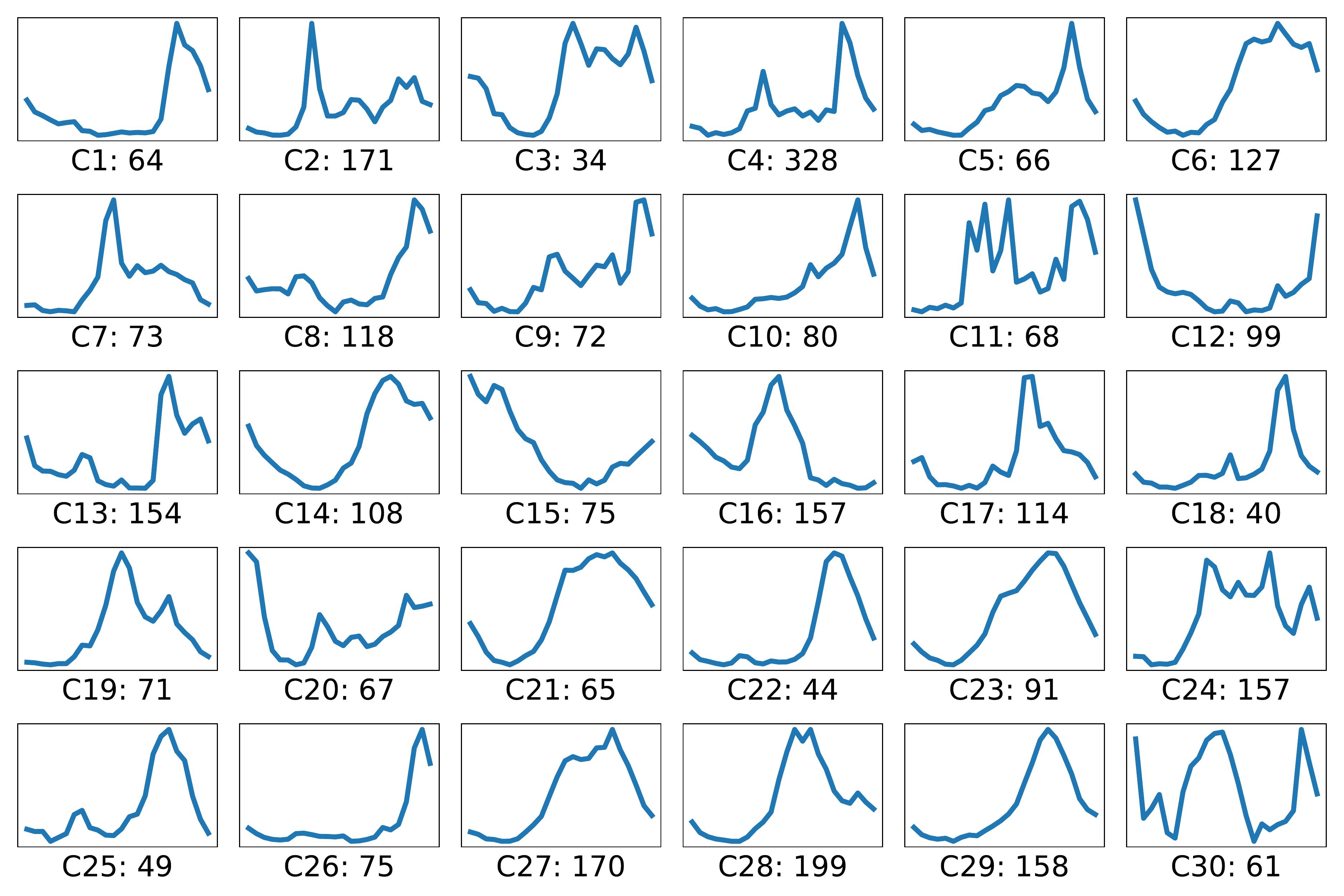}%
\label{fig_first_case}}
\hfil
\subfloat[]{\includegraphics[width=2.1in]{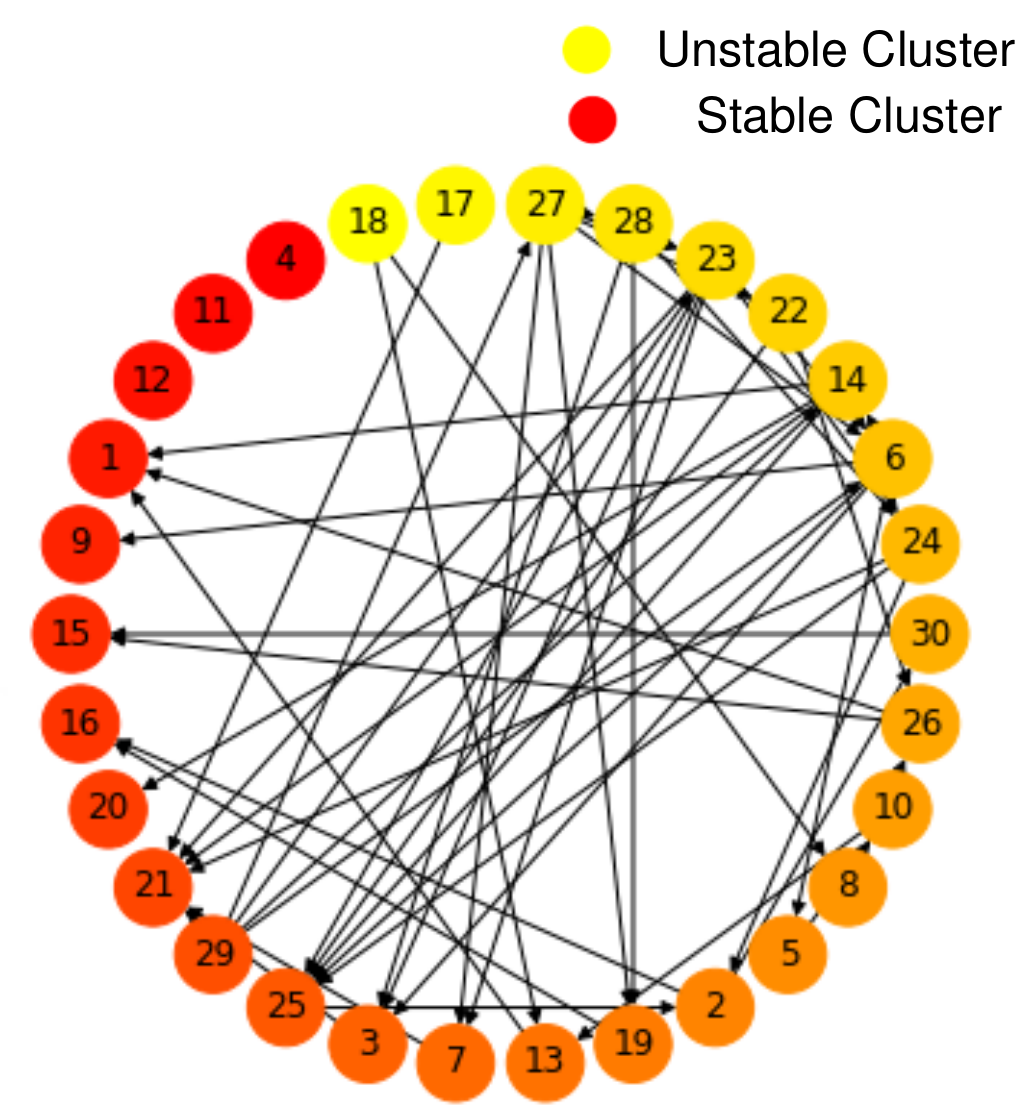}%
\label{fig_second_case}}
\caption{(a) Demand profile of each cluster center, the caption $(Cn: N)$ of each sub figure implies that it is the central profile of cluster $n$, with $N$ users in the cluster (x-axis: time; y-axis: normalized load profile). (b)Disguise trajectories when $\theta=0.01$.}
\label{cluster_result}
\end{figure*}

We conduct the $k$-means clustering for normalized data, and select $k$ to be $30$. Figure \ref{cluster_result}(a) plots the demand profiles for each cluster center. We choose a small $k$ for better illustration of our subsequent ideas toward understanding strategic behaviors. Nevertheless, the central profiles show interesting patterns. For example, cluster $9$ seems more like the classical load pattern with $2$ peaks: one in the noon time and the other at night. However, this cluster only contain $75$ out of the $3,155$ load profiles. We can observe there are many more types of users, including those who are more active at night (cluster $3,12$, and $26$).

\section{Disguising: Strategic Behaviors}\label{Disguising}

Users may strategically change their load profiles for a better price. We need to carefully define ``strategic behaviors''. If a user changes from a cluster center to another cluster center, then its price should be determined by the second one and this pricing scheme is efficient. However, some users may lie on the boundary of two clusters, and hence they can conduct minimal load profile modifications to switch to another cluster, and potentially for a better price. We seek to define such behaviors as disguising. We make the following assumption to conduct the worst case analysis of strategic behaviors:

\vspace{0.1cm}
\noindent $\textbf{Assumption}$: All users know the global information, i.e., the clustering results, including the central profile of each cluster and its corresponding price.
\vspace{0.1cm}

We denote the center (load profile) of cluster $j$ by $c_{j}$, and denote the cluster that user $i$ belongs to by a function $u(i)$. To understand the strategic behavior, we need to examine the effort that user $i$ needs to spend on moving to another cluster $n\neq u(i)$. This effort can be characterized by a scalar $\lambda_{i,n}$. User $i$ can successfully switch to cluster $n$, as long as the following condition holds:
\begin{equation}\label{SwitchCondition}
    ||(1-\lambda_{i,n})\mathbf{d}_{i}+\lambda_{i,n}\mathbf{c}_{n}-\mathbf{c}_{u(i)}||_{1}\geq ||(1-\lambda_{i,n})(\mathbf{d}_{i}-\mathbf{c}_{n})||_{1},
\end{equation}
where the left-hand-side measures the distance between user $i$'s modified profile ($(1-\lambda_{i,n})\mathbf{d}_{i}+\lambda_{i,n}\mathbf{c}_{n}$) and the original cluster center, and the right-hand-side measures the distance between user $i$'s modified profile and the center of cluster $n$. Hence, the minimal effort that user $i$ needs to spend for disguising can be solved by an optimization problem:
\begin{equation}\label{MinMove}
\begin{aligned}
\min_{n\neq u(i)}\,\,& \inf \lambda_{i,n} \\
s.t.\,\,\,\,&||(1-\lambda_{i,n})\mathbf{d}_{i}+\lambda_{i,n}\mathbf{c}_{n}-\mathbf{c}_{u(i)}||_{1}\\
&\ \ \ \ \geq ||(1-\lambda_{i,n})(\mathbf{d}_{i}-\mathbf{c}_{n})||_{1},\\
&p_{n}< p_{u(i)},
\end{aligned}
\end{equation}
where $p_{n}$ represents the price of cluster $n$ (i.e.,$MCI_{n}$).

For notational simplicity, we define
\begin{equation}\label{n}
\begin{aligned}
n_{i}^{*}=\mathop{\arg\min}_{n\in \left \{ 1,...,k \right \},n\neq u(i)} \inf \lambda_{i,n}.
\end{aligned}
\end{equation}

With such characterizations, we use an index $CR$ to better define disguising: for each users $i$, define
\begin{equation}\label{CR}
    CR_{i}=\inf \lambda_{i,n_{i}^{*}}.
\end{equation}

It is worth noting that in the optimization problem (\ref{MinMove}), we intentionally do not choose to minimize the $l_{1}$-norm between modified profile and the original demand profile. Instead, we choose to compare the two profiles' distances to the target cluster center $n$ and define $\lambda_{i,n}$ to reflect the ratio of the two distances. This is our way to `normalize' user's efforts to disguise among all clusters. This establishes the basis for defining $CR_{i}$. Formally,

\vspace{0.1cm}
\noindent $\textbf{Definition 1}$: User $i$ has the ability to disguise if
\begin{equation}\label{theta}
    CR_{i} \leq \theta.
\end{equation}

\noindent $\textbf{Remark}$: In this \textit{parametric} definition, the user's strategic behavior can be measured through a threshold $\theta$. And such threshold cannot be directly observed, but can be inferred through observing the real strategic behaviors, or through understanding how many users have such ability in response to each threshold. This motivates us to characterize the sensitivity zone based on our \textit{parametric} definition of disguising.
\vspace{0.1cm}

We also want to emphasize that by considering the load profile modification based on the normalized profile, we intentionally ignore user's possibility of disguising via reducing energy consumption, which allows us to focus on understanding shift load profile for disguising.

Figure \ref{cluster_result}(b) shows by only strategically adjusting $1\%$ of demand, how the users may disguise and change their clusters. The arrows represent these changes in the clusters. The darker the cluster, the more stable the cluster is, which implies it is more difficult for the users in the cluster to disguise.

\section{Sensitive Zone Characterization}\label{Radius}

Sensitive zones include users that are more likely to modify their load profiles and disguise themselves to belong to other clusters. With the parametric definition of disguising, we characterize the sensitivity zone from two aspects: the number of users in the zone ($N_{n}(\theta)$) and the radius of the stable zone ($r_{n}(\theta)$), which is the complementary set of sensitive zone.

Specifically, we define, for each cluster $n$,
\begin{equation}\label{N}
N_{n}(\theta):=\sum\nolimits_{i,u(i)=n}I(CR_{i}\leq \theta),
\end{equation}
where $I(\cdot )$ is the indicator function.
\begin{equation}\label{radius}
\begin{aligned}
r_{n}(\theta ):=\min_{i}&||\mathbf{c}_{n}-\mathbf{d}_{i}||_{1}\\
s.t.\ \  &CR_{i}\leq \theta, \\
&u(i)=n.
\end{aligned}
\end{equation}

\begin{figure}[!t]
\centering
\includegraphics[width=3.2in]{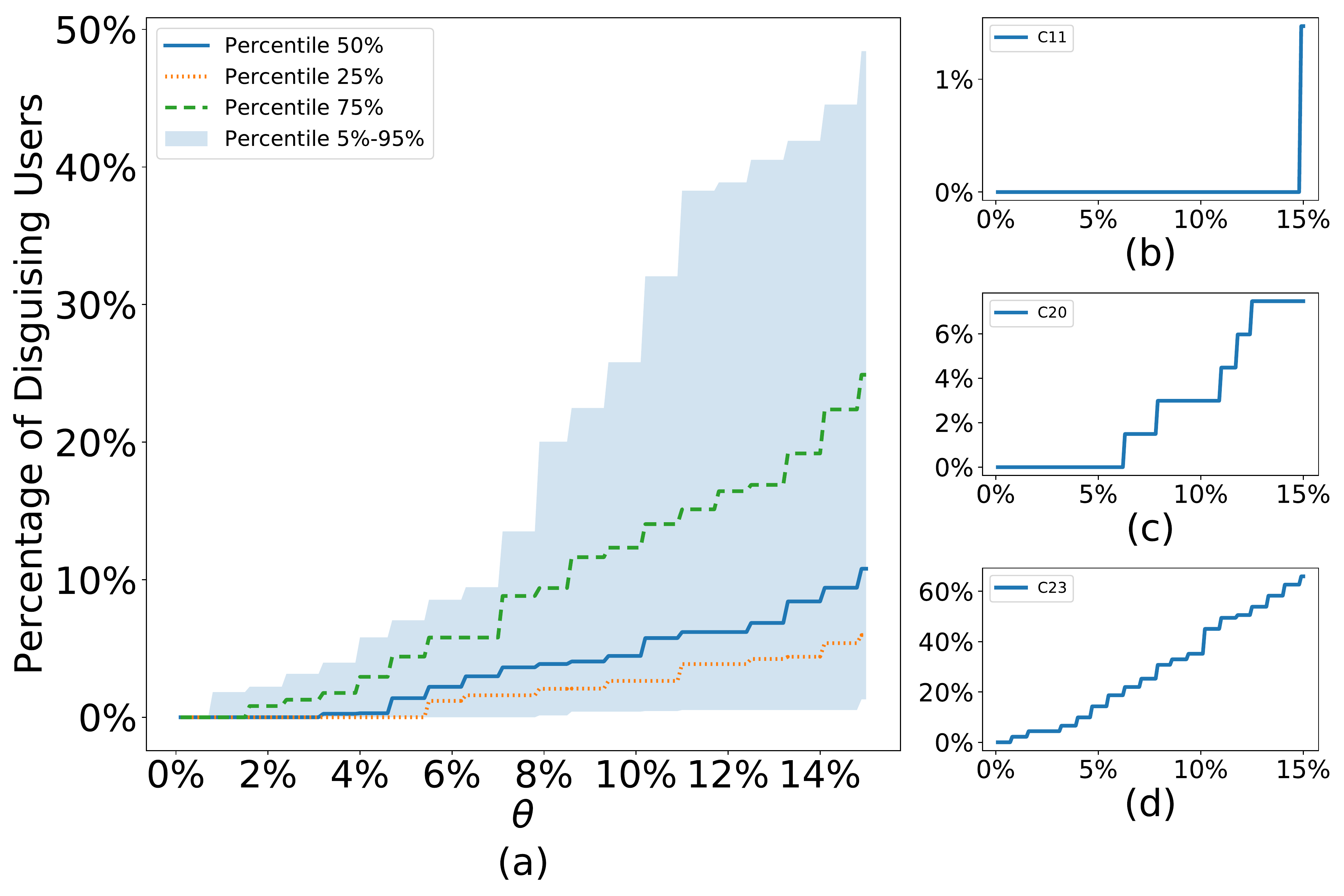}
\caption{Percentage of Strategic Users Evolves with $\theta$.}\vspace{-0.3cm}
\label{NPercent}
\end{figure}

\begin{figure}[!t]
\centering
\includegraphics[width=3.2in]{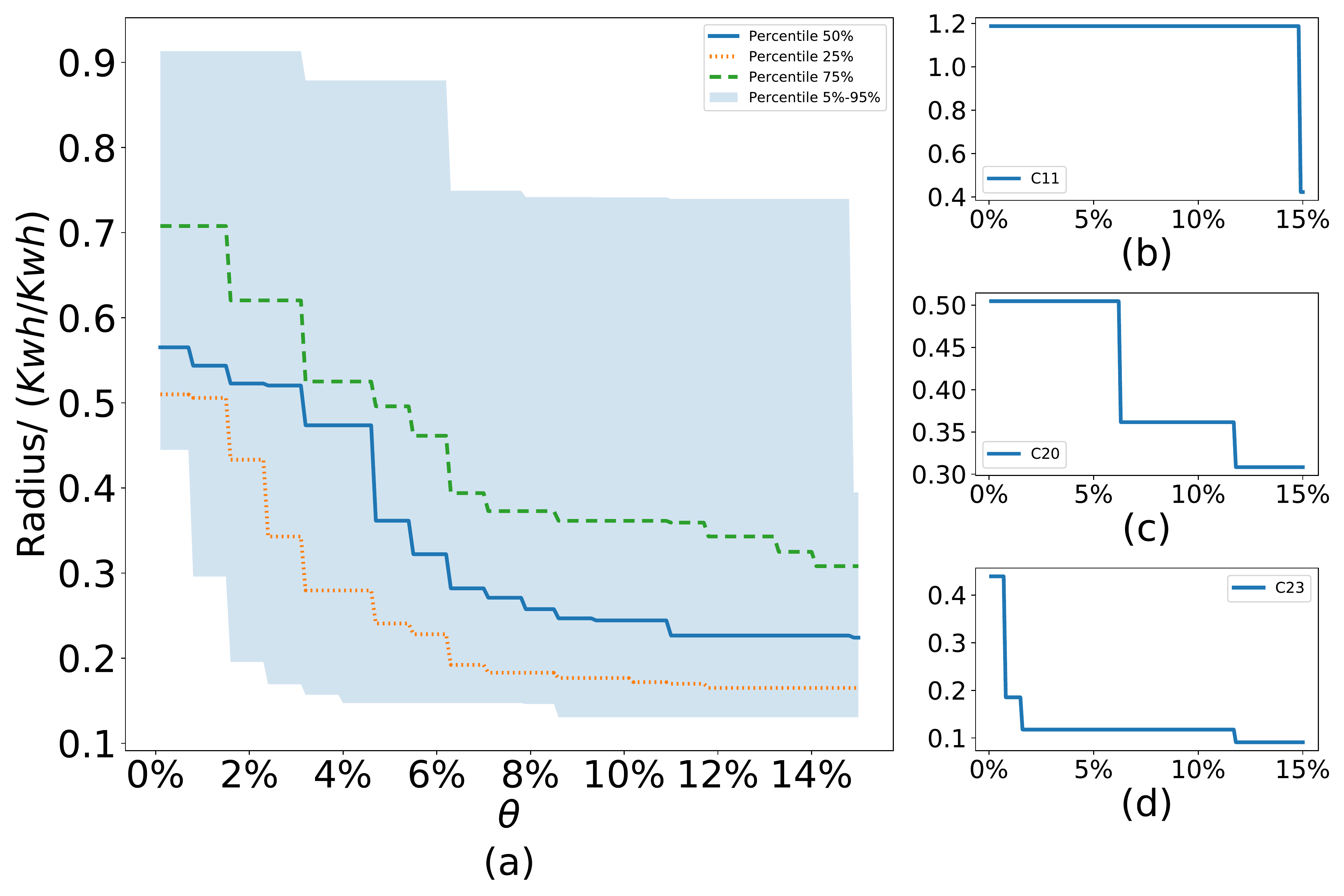}
\caption{Radius ($r_{n}(\theta)$) of Each Cluster Evolves with $\theta$.}\vspace{-0.3cm}
\label{R}
\end{figure}

Figure \ref{NPercent} plots how the percentage of users in the sensitive zone varies with $\theta$. Figure \ref{R}  plots the corresponding radius ($r_{n}(\theta)$) evolving with $\theta$. In the right hand side of the two figures, we illustrate three interesting clusters (cluster $11$, $20$ and $23$) for more observations.

Note that we focus on the profit-seeking strategy of disguising, which is a strategy of obtaining a significant profit by a negligible demand adjustment. Thus, we zoom in the cases when $\theta$ grows from $0\%$ to $15\%$. Figure \ref{NPercent} shows that users can disguise only if they change at least $0.8\%$ of their load profiles. Together with Fig. \ref{R}, the two figures demonstrate heterogeneous stability across clusters. For example, there does not exist the strategic users in cluster $11$ (as shown in Fig. \ref{NPercent}(b)), which is a stable cluster, having a flat radius evolution curve associated with the growth of $\theta$ (as shown in Fig. \ref{R}(b)). Cluster $23$ is an unstable cluster: the number of its strategic users keep growing with $\theta$'s increase while its stable radius keeps on shrinking. 

%We find that although percentage and radius are both increasing with the increase of $\theta$ in all clusters. The increasing rates are neither stable nor monotone. For example, the percentage of strategic users in cluster $8$ (as shown in Fig. \ref{NPercent}(b)) first increases slowly with $\theta$ and after certain point, it starts to rise rapidly with $\theta$. This may correspond to its multiple step changes in the radius evolution (as shown in Fig. \ref{R}(b)). Cluster $33$ (as shown in Fig. \ref{NPercent}(d) and Fig. \ref{R}(d)) also displays such correlation between step changes in radius evolution and non-monotone increasing rate of percentage of strategic users. However, this correlation does not always hold. An exception is cluster $23$, as shown in Fig. \ref{NPercent}(c) and Fig. \ref{R}(c). Such observations illustrate the complex structure of the profile distribution within each cluster: in a high dimensional space, radius is not a good choice for evaluating strategic behaviors. 

%its percentage of strategic users (as shown in Fig. \ref{NPercent}(d)) and radius (as shown in Fig. \ref{R}(d)) demonstrate the step-like change with $\theta$. And cluster $13$ is a cluster with stability between cluster $8$ and cluster $15$. Such observations illustrate the complex structure of the profile distribution within each cluster: in a high dimensional space, radius is not a good choice for evaluating strategic behaviors. }

\section{Vulnerability Analysis}\label{Vulnerability}
To better analyze the incentive of strategic behaviors, we start by conducting the cost benefit analysis of disguising. More precisely, we evaluate the user's potential benefits of disguising by focusing on user's electricity costs and ignoring the user's other utilities (such as discomfort in changing load profile, or the cost of using storage systems for disguising).

\subsection{Cost Benefit Analysis}

%We denote $\mathbf{\widetilde{d}}_{i,n}$ as user $i$'s profile after moving towards cluster $n$ by a proportion of $\mu_{i,n}\in (0,1)$. More precisely,
%\begin{equation}\label{move}
   %\mathbf{\widetilde{d}}_{i,n}=(1-\mu_{i,n})\mathbf{d}_{i}+\mu_{i,n}\mathbf{c}_{n}.
%\end{equation}

With $\mathbf{\widetilde{d}}_{i,n}$, we can evaluate user $i$'s benefit of disguising itself to be a cluster $n$ user. Suppose user $i$ has disguised itself with a disguised profile $\mathbf{\widetilde{d}}_{i,n}$, then its benefit can be measured by the difference in electricity bills $b_{i,n}$:
\begin{equation}\label{benefit}
    b_{i,n}(\mu_{i,n})=p_{u(i)}\sum\nolimits_{t=1}^{24}d_{i}^{t}-p_{n}\sum\nolimits_{t=1}^{24}\widetilde{d}_{i,n}^{t}.
\end{equation}
 %This allows us to evaluate user $i$'s maximal profits of disguising via the following optimization problem:
%\begin{equation}\label{MaxBenefit}
%\begin{aligned}
%\max_{n\neq u(i)}\,\,& \max\left \{ \sup_{\mu_{i,n}}b_{i,n}(\mu_{i,n}),0 \right \} \\
%s.t.\ \ & \mathbf{\widetilde{d}}_{i,n}=(1-\mu_{i,n})\mathbf{d}_{i}+\mu_{i,n}\mathbf{c}_{n},\\
%&||\mathbf{\widetilde{d}}_{i,n}\!-\!\mathbf{c}_{u(i)}||_{1}\geq ||(1\!-\!\mu_{i,n})(\mathbf{d}_{i}\!-\!\mathbf{c}_{n})||_{1},\\
%&0\leq \mu_{i,n}\leq \theta,\\
%&p_{n}<p_{u(i)}.
%\end{aligned}
%\end{equation}
%The second condition guarantees that user $i$ successfully switches itself in cluster $n$; the third condition suggests such switching between clusters is defined as disguising; and the fourth condition implies the possibility of obtaining profits.

%The objective function requires the maximum of $\sup_{\mu_{i,n}}b_{i,n}$ and $0$ to ensure that if all possible strategic behaviors lead to negative benefit, the user would rather stay in its own cluster without changing its profile.

\begin{figure}[!t]
\centering
\includegraphics[width=3.2in]{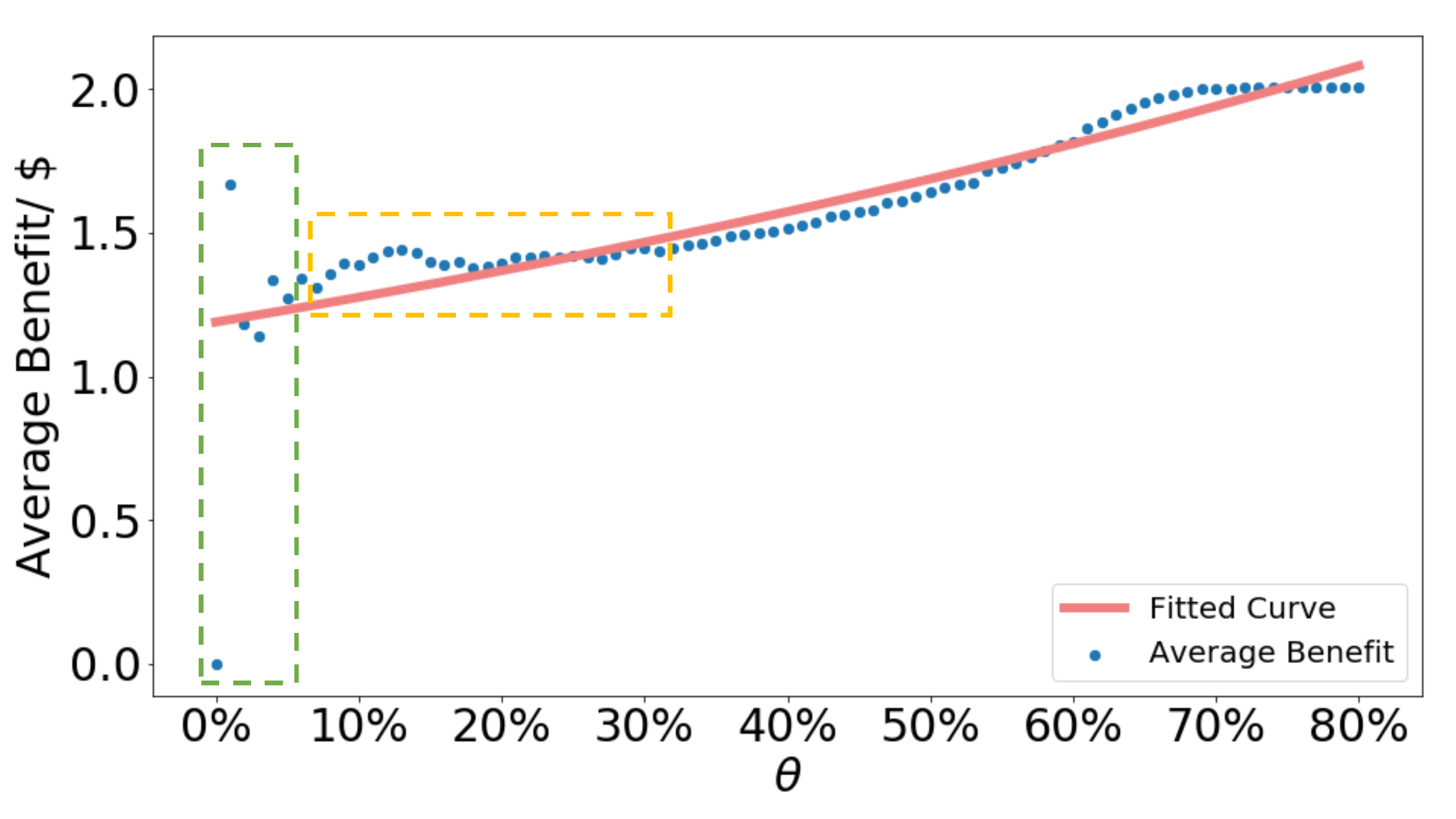}
\caption{Average Benefit with Different $\theta$.}\vspace{-0.3cm}
\label{BenefitExp}
\end{figure}

Note that the value of $\theta$ reflects the difficulty as well as the cost of disguising for a consumer. Whether the consumer will disguise is also contingent on the magnitude of the associated benefit. Hence, we conduct the cost benefit analysis by examining the relation between  $\theta$ and the associated average benefit (as shown in Fig.~\ref{BenefitExp}). 

The non-monotonic $\theta$-benefit relation illustrates the strong incentive for users to disguise. Note that the first impulse of the benefit on the curve in Fig. \ref{BenefitExp} occurs when $\theta$ is only $1\%$, which is even higher than that when $\theta$ is $50\%$. The dramatically high benefit associated with the $1\%$ demand change is the consequence of the disguising strategy's exercise, which further demonstrates the significance of the loopholes induced by the algorithmic limits of the data-driven method. 

Furthermore, the dramatically high profit from the disguising opportunity can even discourage the consumers from fundamentally reshaping their demand profiles, which can truly reduce the system cost. In Fig. \ref{BenefitExp}, we also observe that the marginal benefit remains at a low level when $\theta$ is up to $15\%$. %This implies users in fact have insufficient incentive towards fundamentally reshape their load profiles. 

%Although the trend indicates that a larger $\theta$ often leads to a higher benefit, this relationship is not monotone, especially when $\theta$ is small. The embedded image of Fig. \ref{BenefitExp} zooms in the benefit$-\theta$ relation when $\theta$ falls between $0\%$ to $15\%$. The initial benefit increases dramatically, which corresponds to. This supports our arguments in Section \ref{Clustering} that the data-driven pricing scheme may create loopholes for users to exploit.

%The marginal benefit remains at a low level when $\theta$ is up to $15\%$. This implies users in fact have insufficient incentive towards fundamentally reshape their load profiles. 

%It is straightforward to see that there do exist users who can easily modify their profiles for disguising and obtain large profits. 

\subsection{Vulnerability Analysis}
Though easy to manipulate, the manipulation may improve social welfare. However, in this section, we submit that the manipulation may actually increase the system cost.

\begin{figure}[!t]
\centering
\includegraphics[width=3.2in]{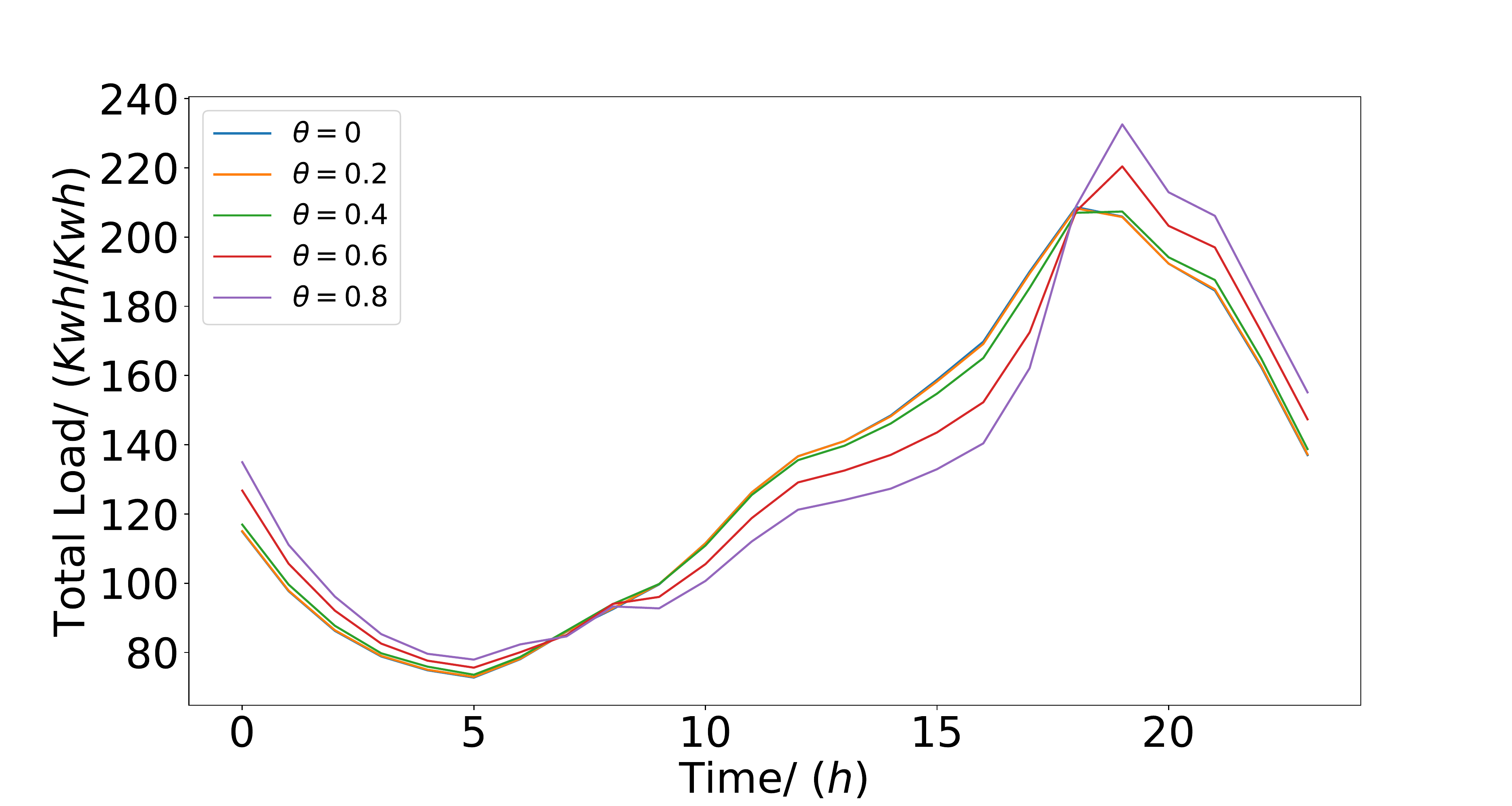}
\caption{Total Load with Different $\theta$.}\vspace{-0.3cm}
\label{Total Load}
\end{figure}

Figure \ref{Total Load} plots the total load in the system with different $\theta$. The system peak increases dramatically for fairly large $\theta$ $(\theta>0.4)$. This is rather counter intuitive as a larger $\theta$ implies that the customers are following the system operator's price signal. The major reason for this market failure is due to the price taker assumption in our model. When $\theta$ is large, the change in the aggregate load profile is enough to affect the real time price and hence the user's marginal system impact. However, a detailed discussion is beyond the scope of this paper. Instead, in this paper, we simply want to identify the possibility that disguising may lead to a higher peak.

\subsection{Generalization}

We have focused on the analysis for the potential benefits of disguising. %We can easily generalize optimization problem (\ref{MaxBenefit}) to accommodate each user's utilities. 
More practically, we may define the utility function $u_{i}(\mathbf{d}_{i,n})$ for user $i$ as follows:
\begin{equation}\label{utility}
   u_{i}(\mathbf{d}_{i,n})=w_{i}(\mathbf{d}_{i,n})-p_{n}\sum\nolimits_{t=1}^{24}d_{i}^{t},
\end{equation}
where $w_{i}(\mathbf{d}_{i,n})$ implies user $i$'s mentally happiness of consuming energy according to pattern $\mathbf{d}_{i,n}$, and the second term implies the electricity cost. Hence, we can accordingly update user $i$'s benefits from disguising to be cluster $n$ user as follows:
\begin{equation}\label{NewBenefit}
    b_{i,n}(u_{i,n})=u_{i}(\mathbf{d}_{i})-u_{i}(\mathbf{d}_{i,n}).
\end{equation}

\vspace{0.2cm}
\noindent $\textbf{Remark}$: One possible form of $w_{i}(\mathbf{\widetilde{d}})$ could be
\begin{equation}\label{UtilityFuc}
    w_{i}(\mathbf{\widetilde{d}}_{i,n})=u_{i}^{\max}-c*||\mathbf{\widetilde{d}}_{i,n}-\mathbf{d}_{i}||_{1}.
\end{equation}
However, it is in general difficult to estimate user $i$'s mentally happiness. Hence, in this paper, we restrict ourselves in understanding the potential benefits of disguising.

%\begin{figure}[!t]
%\centering
%\includegraphics[width=2.8 in]{Peak.pdf}
%\caption{Peak of Total Load with Different $\theta$.}\vspace{-0.3cm}
%\label{Peak}
%\end{figure}

\section{Conclusion}\label{Conclusion}
This paper conducts the vulnerability analysis for data-driven pricing schemes. Using a clustering oriented pricing scheme as an example, we identify strategic behaviors by defining disguising and analyze the impact of disguising on a system level. We submit that disguising can be harmful to the power system operation by leading to a higher peak.

This work can be extended in many ways. For example, we assume each user has full knowledge of the clustering results. It will be interesting to understand the condition when users only know their neighboring clusters' information. We also intend to investigate the following research question: will a user constantly have the ability to easily disguise for profits? This requires a careful examination on the temporal characteristics of the load profiles over a longer time span.

\bibliographystyle{ieeetr}
\bibliography{reference}

\end{document}